\documentclass{article}
\usepackage[T1]{fontenc}
\usepackage{iclr2027_conference,times}

\usepackage{amsmath,amsfonts,bm}

\def\eqref#1{equation~\ref{#1}}

\def\1{\bm{1}}

\DeclareMathAlphabet{\mathsfit}{\encodingdefault}{\sfdefault}{m}{sl}
\SetMathAlphabet{\mathsfit}{bold}{\encodingdefault}{\sfdefault}{bx}{n}

\usepackage{booktabs}
\usepackage{tabularx}
\usepackage{graphicx}
\usepackage{placeins}
\usepackage{xurl}
\usepackage{hyperref}
\usepackage{url}
\hypersetup{hidelinks}

\title{\makebox[\linewidth][s]{Train4Merge:\hfill RL\hfill or\hfill SFT---Which\hfill Yields\hfill Better}\\
Teachers for OPD-Based Model Merging?}
\author{Jingyuan Huang$^{1}$, Zuming Huang$^{2}$, Yucheng Shi$^{3}$, Zhongzhi Li$^{1}$, \\
\textbf{Xiaoming Zhai$^{1}$, Wei Chu$^{2}$, Ninghao Liu$^{4}$}\thanks{Corresponding author.} \\
{\normalfont $^{1}$University of Georgia \quad $^{2}$INF Tech \quad $^{3}$Tencent \quad $^{4}$Hong Kong Polytechnic University}}

\iclrfinalcopy

\begin{document}
\maketitle
\lhead{}
\renewcommand{\headrulewidth}{0pt}

\begin{abstract}
Domain experts trained from a shared checkpoint can transfer their specialized capabilities to a single student through on-policy distillation (OPD). Existing research primarily focuses on improving this merging process, while the algorithms used to train the experts have received limited systematic comparison. We investigate which training algorithm produces teachers better suited to OPD through controlled single-teacher comparisons of supervised fine-tuning (SFT) and reinforcement learning (RL) across Agentic, Reasoning, and Perception. We further examine whether this advantage extends to multi-teacher OPD (MOPD). Teachers and students share the same Qwen3.5-9B initialization, and the two teacher types are compared at similar task performance. Our single-teacher experiments show that RL teachers yield stronger students and higher recovery of teacher performance gains across all three domains. In MOPD, the RL teacher group also yields a student with a higher three-domain mean score and higher overall teacher gain recovery than the SFT teacher group. Our further analysis shows that RL teachers undergo smaller parameter displacements from the shared initialization than SFT teachers. Teacher--student Top-$K$ overlap also increases faster early in OPD or remains consistently higher with RL teachers than with SFT teachers. These findings support the hypothesis that RL teachers' smaller departures from the student's starting point facilitate learning through OPD, resulting in stronger students.
\end{abstract}

\begin{figure}[!ht]
    \centering
    \includegraphics[width=0.90\textwidth]{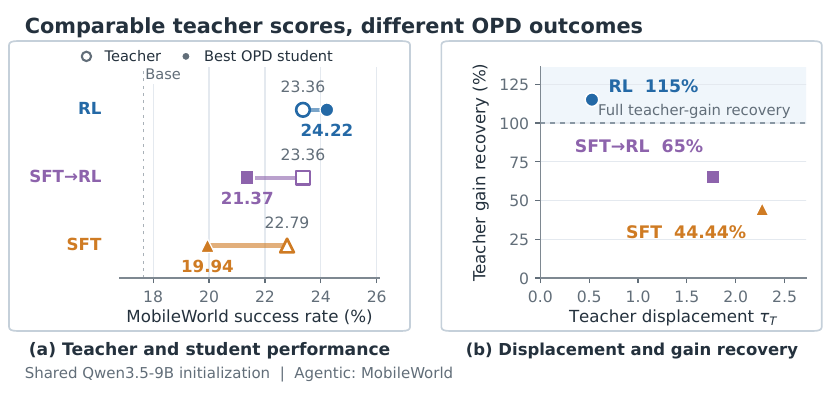}
    \caption{Teacher training and OPD outcomes on MobileWorld.}
    \label{fig:teacher-training-teaser}
\end{figure}
\clearpage

\section{Introduction}
\label{sec:introduction}

On-policy distillation (OPD) trains a student through teacher supervision on student-generated trajectories~\citep{agarwal2024gkd}. A growing line of work uses OPD for model merging in two stages: train domain teachers from a shared checkpoint, then merge their capabilities into a single student through multi-teacher OPD~\citep{ma2026mopd,yan2026maga}. Existing research on OPD and multi-teacher OPD has primarily focused on improving the OPD algorithm used to transfer teachers' capabilities to students, while the choice of algorithm for training the teachers has received less systematic attention. Current pipelines construct teachers using SFT~\citep{yan2026maga,ui_mopd}, RL~\citep{ma2026mopd}, or a combination of these two. Although these strategies can all produce strong domain teachers, teachers with comparable task performance may not be equally effective at transferring their capabilities through OPD. This motivates our central question: \textbf{which post-training algorithm produces teachers better suited to OPD when training starts from the same checkpoint---SFT or RL?}

In multi-teacher OPD, teachers sharing one student can transfer abilities across domains or interfere with one another~\citep{men2026physics}, which could blur the comparison between SFT and RL teachers. For a cleaner comparison, we first study single-teacher OPD, the building block of multi-teacher OPD. We construct SFT and RL teachers with comparable task performance and conduct experiments in Agentic, Reasoning, and Perception under a single-teacher, single-student OPD setting. We then extend the comparison to MOPD, where a group of domain teachers supervises a single student, to examine whether the findings also hold when capabilities from multiple domains are merged.

Our single-teacher results show that students trained with RL teachers consistently outperform those trained with SFT teachers across all three domains. At their best checkpoints, RL-guided students outperform SFT-guided students by 4.27, 0.83, and 0.86 percentage points in Agentic, Reasoning, and Perception, respectively, and recover more of their teachers' performance gains over the base model. In Agentic, this 4.27-point student gap arises from teachers just 0.57 points apart in success rate: the best SFT-guided student recovers only 44.44\% of its teacher's gain, whereas the best RL-guided student reaches 115.00\% and surpasses its teacher. In Reasoning and Perception, the best RL-guided students lead even though their teachers score below the SFT teachers on the metric used to pair them. This overall advantage extends to MOPD: the RL teacher group yields a higher three-domain mean score than the SFT teacher group (60.04\% versus 58.97\%) and higher overall teacher gain recovery (138.27\% versus 84.51\%).

Why do teachers with comparable task performance lead to different student outcomes? Our analysis supports the hypothesis that, under shared initialization, a smaller teacher parameter displacement may help the student align with the teacher’s predictions through OPD and achieve better task performance. Across all three domains, RL teachers exhibit smaller parameter displacements than SFT teachers. In Agentic and Perception, teacher--student Top-$K$ overlap, a measure of agreement between their token predictions, increases faster with RL teachers during the initial stage of OPD; in Reasoning, it remains higher with RL teachers at every measured checkpoint. These parameter displacement and early prediction alignment patterns, together with the stronger performance of RL-guided students, support this explanation of the RL teacher advantage.

A teacher's own score is thus an incomplete guide to its value for OPD. When constructing domain teachers from a shared base model for subsequent OPD, our findings support prioritizing RL. Our main contributions are as follows.
\begin{enumerate}
    \item \textbf{Demonstrating the distillation advantage of RL teachers.} In controlled single-teacher comparisons under shared initialization and comparable teacher performance, the best RL-guided student in each domain outperforms the best SFT-guided student by 0.83--4.27 percentage points and achieves higher teacher gain recovery. MOPD experiments further show higher three-domain mean performance and overall teacher gain recovery with RL teachers.
    \item \textbf{Providing mechanistic evidence for the RL teacher advantage.} Parameter displacement and Top-$K$ alignment dynamics support the hypothesis that RL teachers' smaller departures from the shared initialization facilitate student learning through OPD.
\end{enumerate}

\section{Related Work}
\label{sec:related-work}

\paragraph{Capability integration through model merging and OPD.}
Model merging seeks to integrate specialized capabilities into a single model. Parameter-space approaches combine task vectors relative to a shared base model~\citep{ilharco2023taskarithmetic}, resolve interference between updates~\citep{yadav2023ties}, or sparsify and rescale task-specific changes before merging~\citep{yu2024dare}. OPD provides another route by transferring teacher capabilities through supervision on student-generated trajectories~\citep{agarwal2024gkd}, with Qwen3 demonstrating its use for strong-to-weak transfer following off-policy distillation~\citep{qwen2025qwen3}. Recent work further improves this transfer process through error-aware supervision allocation~\citep{yan2026maga} and curricula over interaction depth~\citep{wang2026tcod}. Other approaches use verifiable correctness feedback to filter, gate, or reweight teacher guidance on student-generated trajectories~\citep{huang2026trust,akhondzadeh2026rgopd,lin2026opdvr,gan2026raopd}. Some works also exploit privileged visual information to improve teacher supervision. Vision-OPD transfers crop-conditioned predictions to a full-image student, while GUI-SD combines visually enriched teacher inputs with entropy-guided token weighting for GUI grounding~\citep{yuan2026visionopd,gui_sd}. These advances focus primarily on how teacher guidance is selected and applied during OPD. The upstream choice of how to train teachers for effective capability merging has received less systematic attention.

\paragraph{Expert construction for subsequent OPD.}
Many works first train multiple teachers and then merge their capabilities into a single model through MOPD, but differ in the algorithms used to train those teachers. Xiaomi's MOPD trains domain experts through RL from a common general SFT checkpoint and initializes the OPD student from that same checkpoint~\citep{ma2026mopd}. MAGA and UI-MOPD instead use SFT to construct platform-specific teachers from a shared checkpoint within each pipeline~\citep{yan2026maga,ui_mopd}. Other pipelines combine these algorithms or use different recipes across domains: DeepSeek-V4 applies SFT followed by GRPO to construct specialists for OPD consolidation~\citep{deepseek2026v4}, while MiMo-V2-Flash uses both SFT- and RL-trained teachers~\citep{xiaomi2026mimov2flash}. Although these pipelines demonstrate several ways to construct experts, they do not establish which training algorithm produces more effective OPD teachers under comparable conditions. This leaves open a central question: when teachers start from the same checkpoint and reach similar task performance, does SFT or RL better support subsequent capability merging?

\paragraph{Teacher compatibility and controlled comparison.}
Mechanistic studies further motivate examining teacher training as a factor in OPD effectiveness. Rethinking OPD relates transfer effectiveness to teacher--student compatibility and the progressive alignment of high-probability tokens~\citep{li2026rethinkingopd}. Studies of parameter dynamics observe small student displacements during multi-teacher OPD~\citep{men2026physics} and concentrated updates with early subspace alignment~\citep{cai2026foresee}; Rethinking OPD II further examines how quickly students absorb teacher supervision~\citep{fu2026rethinkingopd2}. Together, these findings suggest that effective transfer depends on how readily students can acquire teacher capabilities, motivating attention to how those teachers are trained. Our work addresses this upstream choice through controlled single-teacher comparisons: SFT and RL are applied directly to the same initial checkpoint, the resulting teachers have comparable task performance, and both students in each comparison are initialized from that same checkpoint. Across Agentic, Reasoning, and Perception, these comparisons and our MOPD experiments examine which training algorithm better prepares domain experts for subsequent capability merging, complementing work that improves the OPD procedure itself.

\section{Method}
\label{sec:method}

Prior work reports substantially smaller parameter changes during MOPD than during SFT in its experiments~\citep{men2026physics}. This observation suggests that capability transfer requiring large changes to student parameters may be more difficult through OPD. Earlier distillation research also shows that moving the student's initialization closer to the teacher's parameters can improve its ability to match the teacher's predictions~\citep{stanton2021does}. Meanwhile, several studies observe smaller parameter displacements under RL than under SFT~\citep{yuan2025mitigating,cai2026predictability}. These findings raise a question: can the teacher-training algorithm affect subsequent capability merging through OPD by changing the distance between the teacher's parameters and the student's initialization? We therefore hypothesize that, under shared initialization and comparable teacher task performance, RL teachers with smaller parameter displacements may be easier for students to learn from through OPD than SFT teachers, enabling closer prediction alignment and better task performance. To test this hypothesis, we conduct controlled single-teacher OPD and MOPD experiments.

\subsection{Teacher Construction and Distillation}
\label{sec:teacher-construction}

Let $\pi_0=\pi_{\theta_0}$ denote the shared Qwen3.5-9B initialization~\citep{qwen2026qwen35}, and let $\mathcal D_d$ denote the training task pool for domain $d$. We train two teachers in each domain by applying SFT and RL directly to $\pi_0$. SFT learns from task demonstrations, while RL uses group relative policy optimization (GRPO)~\citep{shao2024deepseekmath} to optimize model-generated responses or actions using task rewards. After completing teacher training, we select SFT and RL checkpoints with comparable domain task performance, denoted by $T_{\mathrm{SFT}}^{(d)}$ and $T_{\mathrm{RL}}^{(d)}$. This comparison reduces the influence of teacher task performance when examining how the teacher-training algorithm affects subsequent OPD.

\paragraph{Single-teacher OPD.}
For each selected teacher, we initialize a separate student from $\pi_0$. The student generates responses from the domain training inputs, and the fixed teacher provides supervision on the student-generated prefixes. Writing $a\in\{\mathrm{SFT},\mathrm{RL}\}$ for the teacher-training algorithm, the objective is
\begin{equation}
\mathcal L_{\mathrm{OPD}}^{(d,a)}(\theta;\bar\theta)
=\mathbb E_{(h,z)\sim\mathcal H_{\bar\theta,d}}
\left[\ell_d\!\left(\pi_\theta,T_a^{(d)};h,z\right)\right].
\label{eq:student-opd}
\end{equation}
Here, $\pi_\theta$ is the student being optimized, $\bar\theta$ denotes the student parameters used to collect the current rollouts, and $h$ contains the input and a generated token prefix. The token $z$ is the student-generated continuation at that position. The distribution $\mathcal H_{\bar\theta,d}$ weights valid response-token positions in those rollouts equally, and $\ell_d$ denotes the token-level distillation loss that guides the student to learn the teacher's predictions on student-generated prefixes. The student continues to generate new rollouts during training, with supervision provided by the fixed teacher.

Within each domain, the two student groups use the same training inputs and OPD configuration. The teacher-training algorithm is the variable under comparison. We denote the resulting students by $S_{\leftarrow a}^{(d)}$.

\paragraph{Multi-teacher OPD.}
We next extend the comparison to merging capabilities from all three domains into one student. The three SFT teachers supervise one student, and the three RL teachers supervise another; both students start from $\pi_0$. Each training example is routed to the teacher for its domain. For domain-routed MOPD, the objective takes the mixture form
\begin{equation}
\mathcal L_{\mathrm{MOPD}}^{(a)}(\theta;\bar\theta)
=\sum_{d\in\mathcal G}w_d\,
\mathbb E_{(h,z)\sim\mathcal H_{\bar\theta,d}}
\left[\ell_{\mathrm{MOPD},d}\!\left(\pi_\theta,T_a^{(d)};h,z\right)\right],
\qquad \sum_{d\in\mathcal G}w_d=1.
\label{eq:student-mopd}
\end{equation}
Here, $\mathcal G$ is the set of the three domains, $w_d\geq0$ denotes the weight of domain $d$ in the training mixture, and $\ell_{\mathrm{MOPD},d}$ is its token-level distillation loss in the MOPD run. The student parameters $\theta$ are shared across domains, while the supervising teacher is determined by the input domain. This formulation makes the comparison explicit: single-teacher OPD examines each teacher separately, whereas MOPD examines how each teacher group supports capability merging in a shared student. Teacher selection criteria and evaluation metrics are described in Section~\ref{sec:experimental-setup}; training details are provided in Appendix~\ref{app:experimental-details}.

\subsection{Training Data}
\label{sec:training-data}

Within each domain, teacher training and single-teacher OPD draw on the same training task pool. SFT uses its demonstration targets, RL uses its task rewards, and OPD uses teacher supervision on student-generated responses.

\paragraph{Agentic.}
We use the publicly released mobile training data from UI-MOPD~\citep{ui_mopd}\footnote{\url{https://github.com/EliSpectre/UI-MOPD/tree/main/data}}. SFT learns from demonstration actions, whereas RL optimizes model-generated actions using task rewards.

\paragraph{Reasoning.}
We use Geometry3K (Geo3K)~\citep{lu2021intergps} as the common source of mathematical training data. Both teachers use the Geo3K training problems in the Geometry3K subset of NVIDIA's Nemotron-Image-Training-v3~\citep{nvidia_geo3k}. The SFT teacher uses its complete chain-of-thought annotations as supervision targets, whereas the RL teacher uses the original Geo3K answer labels as reward references.

\paragraph{Perception.}
We use the publicly released GUI grounding training data from GUI-SD~\citep{gui_sd}\footnote{\url{https://huggingface.co/datasets/yankie123/GUI-SD-data}}. Each example consists of an interface screenshot, a target description, and a location annotation. SFT learns from demonstration locations, whereas RL optimizes predicted locations using task rewards. Dataset sizes and preprocessing details are provided in Appendix~\ref{app:experimental-details}.

\section{Experimental Setup}
\label{sec:experimental-setup}

We select SFT and RL teachers with comparable task performance within each domain and initialize all students from the same Qwen3.5-9B checkpoint. The selected teachers are reused in both single-teacher OPD and MOPD. Table~\ref{tab:initial-teacher-performance} summarizes their performance alongside the initial student under the domain criteria defined below.

\begin{table}[htbp]
    \caption{Initial student and teacher performance (\%).}
    \label{tab:initial-teacher-performance}
    \centering
    \small
    \begin{tabular*}{\linewidth}{@{\extracolsep{\fill}}lccc@{}}
        \toprule
        Model & Agentic & Reasoning & Perception \\
        \midrule
        Initial student (Qwen3.5-9B) & 17.66 & 79.70 & 65.19 \\
        SFT teachers & 22.79 & 88.52 & 68.24 \\
        RL teachers & 23.36 & 84.19 & 67.71 \\
        \bottomrule
    \end{tabular*}
\end{table}

\subsection{Evaluation Benchmarks and Metrics}
\label{sec:evaluation}

We evaluate Agentic performance on the MobileWorld GUI-only task set over three rounds, with 117 tasks per round. The reported mean success rate is the total number of successes divided by 351. For Reasoning, we report answer accuracy on Geo3K, MathVista, MathVision, and MathVerse in the single-teacher experiments, and use Geo3K as the domain score in MOPD. Perception is evaluated on six GUI grounding benchmarks. The benchmarks and their task metrics are summarized in Table~\ref{tab:benchmarks}.

\begin{table}[htbp]
    \caption{Evaluation benchmarks and metrics.}
    \label{tab:benchmarks}
    \centering
    \begin{tabularx}{\linewidth}{@{}>{\raggedright\arraybackslash}p{0.14\linewidth} >{\raggedright\arraybackslash}X >{\raggedright\arraybackslash}p{0.17\linewidth}@{}}
        \toprule
        Domain & Evaluation benchmarks & Task metric \\
        \midrule
        Agentic & MobileWorld~\citep{mobileworld} & Task success rate \\
        \addlinespace
        Reasoning & Geo3K test set~\citep{lu2021intergps}, MathVista~\citep{lu2024mathvista}, MathVision~\citep{wang2024mathvision}, MathVerse~\citep{zhang2024mathverse} & Answer accuracy \\
        \addlinespace
        Perception & ScreenSpot-v2~\citep{screenspot_v2}, ScreenSpot-Pro~\citep{screenspot_pro}, UI-Vision~\citep{uivision}, MMBench-GUI L2~\citep{mmbench_gui}, OSWorld-G, OSWorld-G-Refine~\citep{osworld_g} & Grounding accuracy \\
        \bottomrule
    \end{tabularx}
\end{table}

\paragraph{Evaluation criteria.}
Teacher checkpoints are selected using MobileWorld mean success rate for Agentic, Geo3K accuracy for Reasoning, and the unweighted mean accuracy across the six benchmarks for Perception. Let $J_d(\cdot)$ denote the corresponding criterion in domain $d$, with higher scores indicating better performance. We use two complementary metrics to assess distillation effectiveness.

\paragraph{Domain task performance.}
We compare student performance under $J_d$ throughout training, using the initial student and both teachers as references. For each single-teacher run, we select the student checkpoint with the highest $J_d$ and report all benchmark scores from that checkpoint. Per-benchmark results for all student checkpoints are provided in Appendix~\ref{app:checkpoint-results}.

\paragraph{Teacher performance gain recovery rate.}
We measure the fraction of a teacher's post-training gain recovered by its student as
\begin{equation}
\operatorname{Recovery}_{d,a}
=\frac{J_d\!\left(S_{\leftarrow a}^{(d)}\right)-J_d(\pi_0)}
{J_d\!\left(T_a^{(d)}\right)-J_d(\pi_0)}\times100\%.
\label{eq:gain-recovery}
\end{equation}
The numerator is the student's performance gain over the initial model, and the denominator is the corresponding teacher's gain. Values above 100\% indicate that the student surpasses its teacher. Recovery is defined only when the teacher outperforms the initial model; otherwise, we report N/A. For single-teacher OPD, we report recovery at each group's best checkpoint under $J_d$.

For MOPD, we select each group's best checkpoint by the unweighted mean of its Agentic, Reasoning, and Perception scores and report all domain scores from that checkpoint. Overall teacher gain recovery applies the same calculation to the three-domain means of the student, its teacher group, and the initial model. The teacher-group mean uses each teacher's score in its own domain.

The initial model, teachers, and students use the same evaluation procedure and scoring rules on each benchmark. Evaluation splits, decoding settings, and aggregation details are provided in Appendix~\ref{app:experimental-details}.

\subsection{Diagnostic Metrics}
\label{sec:diagnostic-metrics}

To examine the hypothesis introduced in Section~\ref{sec:method}, we measure parameter displacement from the shared initialization and teacher--student prediction alignment during OPD. The first characterizes how far each teacher departs from the student's starting point; the second tracks how the student's predictions align with its teacher during training.

\paragraph{Parameter displacement.}
Let $\theta_T$ denote the teacher parameters and $\theta_t$ the student parameters after $t$ OPD updates. Their displacements from the shared initialization $\theta_0$ are
\begin{equation}
\tau_T=\|\theta_T-\theta_0\|_2,
\qquad r_t=\|\theta_t-\theta_0\|_2.
\label{eq:parameter-displacement}
\end{equation}
Here, $\|\cdot\|_2$ is the $L_2$ norm, and $\tau_T$ and $r_t$ measure the absolute parameter displacements of the teacher and student, respectively. Both norms are computed over the same 760 non-MTP parameter tensors; for reference, $\|\theta_0\|_2=1796.80$. Smaller values indicate less parameter change from the shared initialization.

\paragraph{Top-$K$ overlap.}
For a prefix $h$, let $S_K(\theta,h)$ be the set of the $K$ highest-probability next tokens under parameters $\theta$. At diagnostic point $t$, we compute
\begin{equation}
O_K(t)=\mathbb E_{h\sim\mathcal H_t}
\left[\frac{|S_K(\theta_t,h)\cap S_K(\theta_T,h)|}{K}\right],
\label{eq:topk-overlap}
\end{equation}
Here, $\cap$ denotes set intersection, $|\cdot|$ denotes the number of tokens in a set, and $\mathcal H_t$ weights valid response-token positions in student-generated rollouts at diagnostic point $t$ equally. Larger overlap indicates closer agreement between teacher and student predictions. We use $K=20$ and compare both overlap levels and their changes during training to characterize prediction alignment.

\section{Experimental Results}
\label{sec:results-analysis}

We first compare SFT and RL teachers through single-teacher OPD across Agentic, Reasoning, and Perception, then examine whether the findings extend to MOPD. Figure~\ref{fig:single-teacher-results} shows the single-teacher learning curves, with both teachers and the base model as references. For single-teacher OPD, teacher gain recovery is reported for each group's best checkpoint. Per-benchmark results for the single-teacher experiments are provided in Appendix~\ref{app:checkpoint-results}.

\subsection{Single-Teacher OPD}
\label{sec:results-single-teacher}

\begin{figure}[!htbp]
    \centering
    \includegraphics[width=\linewidth]{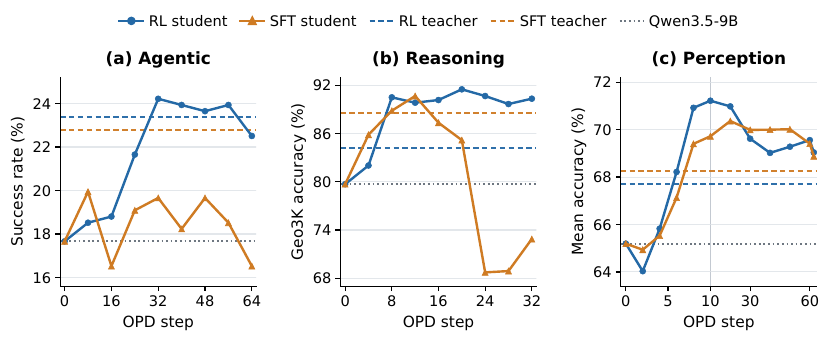}
    \caption{Single-teacher OPD results: (a) Agentic, (b) Reasoning, and (c) Perception. The Perception axis is expanded for the first ten steps.}
    \label{fig:single-teacher-results}
\end{figure}

Across all three domains, RL teachers yield stronger students and higher teacher gain recovery in single-teacher OPD (Figure~\ref{fig:single-teacher-results}). The contrast is most pronounced in Agentic: although the SFT and RL teachers achieve similar success rates of 22.79\% and 23.36\%, their best students achieve 19.94\% and 24.22\%, respectively. The SFT-guided student recovers only 44.44\% of its teacher's performance gain, whereas the RL-guided student recovers 115.00\%, surpassing its teacher. In Reasoning, the RL teacher scores below the SFT teacher on Geo3K (84.19\% versus 88.52\%), yet its student achieves higher accuracy (91.51\% versus 90.68\%) and gain recovery (262.96\% versus 124.53\%). Perception follows the same pattern: despite a lower teacher mean (67.71\% versus 68.24\%), the RL-guided student achieves a higher mean accuracy (71.22\% versus 70.36\%) and gain recovery (239.09\% versus 169.32\%). These results consistently favor RL teachers for OPD. The limited recovery from the SFT teacher on MobileWorld, compared with full recovery in Reasoning and Perception, further suggests that teacher selection may be particularly important for Agentic tasks.

\subsection{Multi-Teacher OPD}
\label{sec:results-mopd}

\begin{figure}[!htbp]
    \centering
    \includegraphics[width=0.75\linewidth]{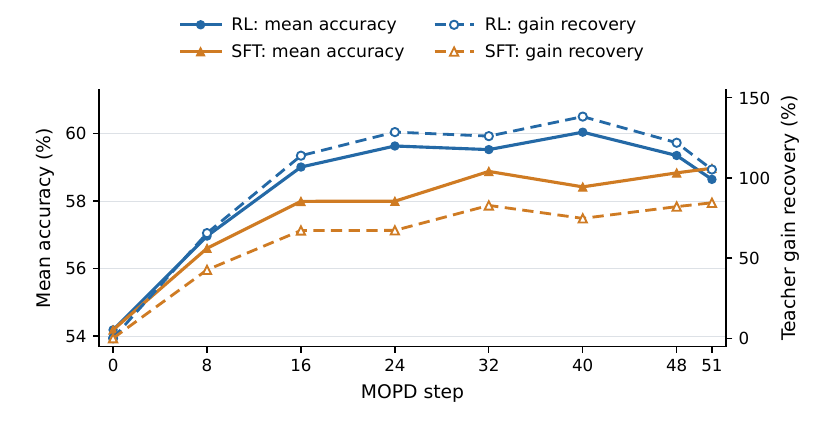}
    \caption{MOPD learning curves. Solid lines show mean accuracy (left axis); dashed lines show teacher gain recovery (right axis).}
    \label{fig:mopd-training-dynamics}
\end{figure}

Having compared the domain teachers individually, we next use the same teachers in MOPD to examine whether the RL advantage persists when their capabilities are merged into a single student. Figure~\ref{fig:mopd-training-dynamics} traces the three-domain mean accuracy and overall teacher gain recovery throughout training.

The RL teacher group again yields better overall results: its best student achieves a three-domain mean score of 60.04\%, compared with 58.97\% for the SFT group, and overall teacher gain recovery of 138.27\%, compared with 84.51\%. Table~\ref{tab:mopd-results} reports the domain teachers' scores alongside those of each group's best MOPD student.

\begin{table}[!htbp]
    \caption{Teacher and MOPD student performance (\%). Parentheses show teacher gain recovery.}
    \label{tab:mopd-results}
    \centering
    \small
    \begin{tabular*}{\linewidth}{@{\extracolsep{\fill}}lcccc@{}}
        \toprule
        Model & Agentic & Reasoning & Perception & Mean \\
        \midrule
        Qwen3.5-9B & 17.66 & 79.70 & 65.19 & 54.19 \\
        SFT teachers & 22.79 & 88.52 & 68.24 & 59.85 \\
        RL teachers & 23.36 & 84.19 & 67.71 & 58.42 \\
        \midrule
        \raisebox{3.2pt}[0pt][0pt]{SFT-guided student} & \shortstack{19.94\\{\scriptsize (+44.44\%)}} & \shortstack{\textbf{89.68}\\{\scriptsize (+113.21\%)}} & \shortstack{67.30\\{\scriptsize (+68.89\%)}} & \shortstack{58.97\\{\scriptsize (+84.51\%)}} \\
        \raisebox{3.2pt}[0pt][0pt]{RL-guided student} & \shortstack{\textbf{23.93}\\{\scriptsize \textbf{(+110.00\%)}}} & \shortstack{86.52\\{\scriptsize \textbf{(+151.85\%)}}} & \shortstack{\textbf{69.68}\\{\scriptsize \textbf{(+177.98\%)}}} & \shortstack{\textbf{60.04}\\{\scriptsize \textbf{(+138.27\%)}}} \\
        \bottomrule
    \end{tabular*}
\end{table}

The RL-guided student performs better in Agentic and Perception, whereas the SFT-guided student scores higher in Reasoning. The gap in Reasoning may partly reflect the SFT teacher's substantially stronger performance in this domain. Nevertheless, the RL teacher group produces a stronger student overall and enables greater recovery of teacher gains, despite having a lower three-domain teacher mean. Together, the single-teacher and MOPD results show that RL teachers support more effective learning for OPD-based Merging.

\section{Mechanistic Analysis of the RL Teacher Advantage}
\label{sec:mechanistic-analysis}

Why are RL teachers more effective for OPD than SFT teachers with comparable task performance? Prior studies report smaller overall parameter displacements under RL than under SFT \citep{yuan2025mitigating,cai2026predictability,ren2026enough}. Meanwhile, prior work reports that models undergo substantially smaller parameter changes during MOPD than during SFT \citep{men2026physics}. We therefore hypothesize that the relatively small parameter changes during OPD favor learning from teachers with smaller displacements from the shared initialization, enabling students to align more readily with teacher predictions and achieve better task performance. We examine this hypothesis first in parameter space and then in output space.

\subsection{Teacher Displacement and OPD Performance}
\label{sec:teacher-parameter-progress}

\begin{figure}[!htbp]
    \centering
    \includegraphics[width=\textwidth]{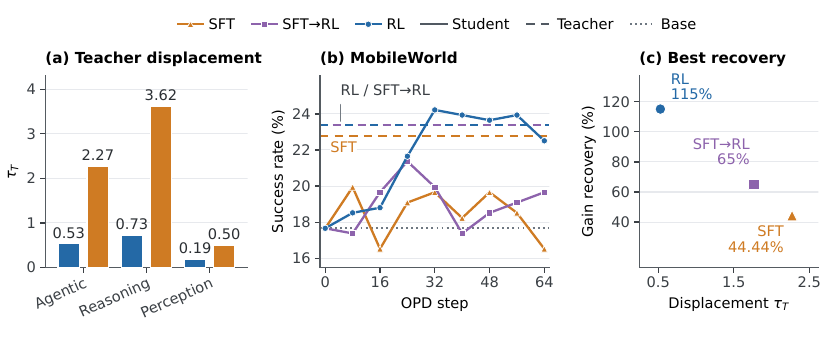}
    \caption{\textbf{Teacher displacement and OPD performance.} (a) Comparison across domains. (b) Agentic accuracy trajectories. (c) Gain recovery at each group's best student checkpoint.}
    \label{fig:teacher-parameter-progress}
    \label{fig:agentic-teacher-routes}
\end{figure}

Figure~\ref{fig:teacher-parameter-progress}(a) shows that, in all three domains, the RL teacher departs less from the shared initialization than the SFT teacher: its displacement $\tau_T$ is $0.23$, $0.20$, and $0.37$ times that of the SFT teacher in Agentic, Reasoning, and Perception, respectively. The performance-matched RL teachers thus reach comparable task performance with much smaller parameter changes, consistent with prior studies. Perception has the smallest difference in teacher displacement magnitudes and a small gap of $0.86$ percentage points between the best students. This observation is consistent with our hypothesis that a teacher's displacement from the shared initialization may affect how readily its capabilities transfer to the student through OPD.

\begin{samepage}
To further examine the relationship between teacher parameter displacement and the performance of the resulting students, we compare three teacher-training paths in Agentic: SFT alone, SFT followed by RL (SFT$\rightarrow$RL), and RL alone. All three paths begin from the shared initialization, and we select teachers with similar task performance. Each OPD student is initialized from the original Qwen3.5-9B checkpoint and trained with its corresponding teacher.\par
\end{samepage}

The three teachers achieve success rates of $22.79\%$, $23.36\%$, and $23.36\%$, respectively, yet their parameter displacements differ substantially: $2.271$, $1.765$, and $0.528$. Their best students achieve success rates of $19.94\%$, $21.37\%$, and $24.22\%$, with teacher gain recovery of $44.44\%$, $65.00\%$, and $115.00\%$ (Figure~\ref{fig:agentic-teacher-routes}(b--c)). Across these three paths, larger teacher displacements are associated with lower recovery of teacher performance gains.

The comparison between the SFT$\rightarrow$RL and direct-RL teachers makes this relationship particularly clear. Both teachers achieve the same success rate of $23.36\%$, but the direct-RL teacher has a much smaller displacement and produces a student whose best success rate is $2.85$ percentage points higher. Thus, even when both teachers have undergone RL and achieve the same task score, their effectiveness for OPD can differ. These results further support our hypothesis that teachers closer to the student's shared initialization are easier to learn from through OPD, helping explain the stronger student produced by the direct-RL teacher.

\subsection{Teacher--Student Prediction Alignment}
\label{sec:parameter-prediction-alignment}

\begin{figure}[!htbp]
    \centering
    \includegraphics[width=\textwidth]{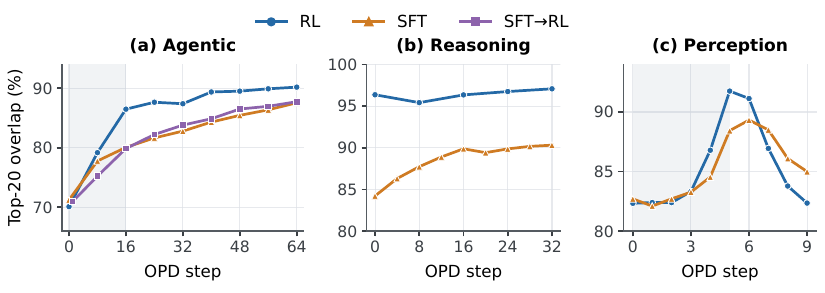}
    \caption{\textbf{Teacher--student Top-20 overlap on student rollouts.} Step~0 denotes the base student before its first update. Shading marks the early steps compared in the text.}
    \label{fig:prediction-alignment-dynamics}
\end{figure}

We next examine how the smaller parameter displacement of RL teachers relates to their students' prediction alignment during OPD. Our hypothesis is that teachers closer to the shared initialization are easier for students to follow: their output agreement can increase faster or remain higher throughout training, helping students achieve better task performance. We measure this agreement using Top-20 overlap on student rollouts across all three domains (Figure~\ref{fig:prediction-alignment-dynamics}).

In both Agentic and Perception, the RL group exhibits substantially larger early increases in overlap than the SFT group. In Agentic, overlap rises by $16.37$ percentage points (pp) over the first 16 steps, compared with $8.81$~pp for SFT. In Perception, the increase over the first five steps is $9.42$~pp for RL and $5.73$~pp for SFT. Thus, although the RL group starts with slightly lower overlap in both domains, its student aligns with the teacher more rapidly early in OPD. In Reasoning, the RL group already has higher overlap at initialization ($96.35\%$ versus $84.21\%$) and remains higher at every measured checkpoint, reaching $97.07\%$ versus $90.31\%$ at step~32.

\FloatBarrier
To further examine the relationship between teacher displacement and top-k overlap increment during OPD, we compare the SFT, SFT$\rightarrow$RL, and direct-RL trajectories in Agentic. Both the direct-RL and SFT$\rightarrow$RL groups begin with lower observed overlap than the SFT group, yet achieve larger overall increases and higher final overlap. Their total gains are $20.07$ and $16.77$ percentage points, respectively, compared with $16.31$ for SFT; their final overlaps reach $90.17\%$ and $87.73\%$, compared with $87.55\%$ for SFT. The direct-RL teacher, which has the smallest parameter displacement, yields the largest overlap gain and the strongest student, while SFT$\rightarrow$RL falls between direct RL and SFT on all three measures (Figure~\ref{fig:agentic-teacher-routes}). Prior work links increasing teacher--student overlap with successful distillation and highlights the importance of early alignment \citep{li2026rethinkingopd}. These output-alignment patterns are consistent with the smaller teacher displacements and larger student gains observed in our experiments. Together, these findings support our hypothesis that teachers closer to the shared initialization are easier for students to follow through OPD, helping explain why RL teachers yield stronger students than comparably performing SFT teachers.
\FloatBarrier

\FloatBarrier

\section{Conclusion}
\label{sec:conclusion}

Our experiments show that the choice of teacher-training algorithm substantially affects the performance of subsequent model merging. Under shared initialization and comparable teacher task performance, RL teachers yield stronger students and higher teacher gain recovery across Agentic, Reasoning, and Perception. The difference is most pronounced in Agentic setting, suggesting that teacher choice may be especially important in this domain. These findings support prioritizing RL when constructing teachers for OPD. Our analyses of parameter displacement and teacher--student Top-$K$ overlap help explain this advantage. RL teachers remain closer to the shared initialization, while their students show faster early prediction alignment or consistently higher output overlap. Together with the student performance results, these observations support our hypothesis that teachers closer to the shared initialization are easier for students to follow through OPD, enabling them to learn more effectively from teacher supervision.

\bibliography{iclr2027_conference}
\bibliographystyle{iclr2027_conference}

\clearpage
\appendix
\section{Additional Details for the Mechanistic Analysis}
\label{app:mechanistic-details}
\label{app:additional-analysis}

\paragraph{Parameter displacement.}
All parameter comparisons use the shared initialization and the same parameter subset. Across all three domains, RL teachers have smaller parameter displacements than SFT teachers.

Table~\ref{tab:student-displacement} reports student displacements $r_t$ in Agentic, Reasoning, and Perception. At the early checkpoints shown, students guided by RL and SFT teachers have similar displacement magnitudes, despite the larger displacements of the SFT teachers.

\begin{table}[htbp]
\centering
\small
\caption{Student parameter displacement $r_t$ and teacher displacement $\tau_T$.}
\label{tab:student-displacement}
\setlength{\tabcolsep}{6pt}
\begin{tabular}{@{}ccc@{\hspace{16pt}}ccc@{\hspace{16pt}}ccc@{}}
\toprule
\multicolumn{3}{c}{Agentic} & \multicolumn{3}{c}{Reasoning} & \multicolumn{3}{c}{Perception} \\
\cmidrule(r{8pt}){1-3}\cmidrule(lr{8pt}){4-6}\cmidrule(l){7-9}
Step & RL & SFT & Step & RL & SFT & Step & RL & SFT \\
\midrule
0 & 0 & 0 & 0 & 0 & 0 & 0 & 0 & 0 \\
8 & 0.209 & 0.197 & 8 & 0.481 & 0.454 & 10 & 0.374 & 0.384 \\
16 & 0.380 & 0.383 & 16 & 0.715 & 0.759 & 20 & 0.629 & 0.726 \\
24 & 0.505 & 0.588 & 24 & 0.828 & 0.929 & 30 & 0.760 & 0.864 \\
32 & 0.579 & 0.714 & 32 & 0.901 & 1.043 & 40 & 0.855 & 0.969 \\
40 & 0.618 & 0.814 & & & & 50 & 0.942 & 1.055 \\
48 & 0.640 & 0.908 & & & & 60 & 1.030 & 1.119 \\
56 & 0.656 & 0.993 & & & & 62 & 1.047 & 1.133 \\
64 & 0.668 & 1.066 & & & & & & \\
\midrule
$\tau_T$ & 0.528 & 2.271 & $\tau_T$ & 0.728 & 3.619 & $\tau_T$ & 0.188 & 0.502 \\
\bottomrule
\end{tabular}
\end{table}

\FloatBarrier

\section{Experimental Details}
\label{app:experimental-details}

\subsection{Training Data}

Agentic uses 25,904 step-level examples from 2,640 trajectories in the UI-MOPD Uni-GUI-OpenMobile release, retaining its original split and input contexts. Reasoning draws inputs from 2,078 Geo3K training problems aligned with NVIDIA's Nemotron-Image-Training-v3 Geometry3K subset. Perception uses 6,990 GUI-SD training examples. In each domain, both student groups use the same training input pool. OPD uses the original prompts and images without reference answers or additional teacher inputs.

\subsection{Teacher Training Configurations}
\label{app:teacher-training}

Table~\ref{tab:teacher-configurations} summarizes the teacher settings. All teachers start from Qwen3.5-9B and use the training data described above. SFT minimizes token-level cross-entropy on demonstration responses. RL uses GRPO with sequence-level task rewards and no entropy or length terms; rollouts use temperature 1.0 (0.9 in Perception), top-$p$ 1.0, and no top-$k$ truncation. Optimizer betas are $(0.9,0.999)$, or $(0.9,0.95)$ for the Reasoning SFT teacher, and the gradient clipping norm is 1. No teacher uses LoRA. All teachers keep the visual encoder and merger frozen during training.

\begin{table}[htbp]
\centering
\footnotesize
\setlength{\tabcolsep}{4.5pt}
\caption{Teacher training settings. Batch: demonstrations (SFT) or prompts $\times$ sampled responses (RL) per update. Max tokens: total (SFT) or prompt/response (RL). Learning rates are configured values.}
\label{tab:teacher-configurations}
\begin{tabular}{@{}lcccccc@{}}
\toprule
 & \multicolumn{2}{c}{Agentic} & \multicolumn{2}{c}{Reasoning} & \multicolumn{2}{c}{Perception} \\
\cmidrule(lr){2-3}\cmidrule(lr){4-5}\cmidrule(l){6-7}
Parameter & SFT & RL & SFT & RL & SFT & RL \\
\midrule
Optimizer & AdamW & AdamW & AdamW & AdamW & AdamW & AdamW \\
Learning rate & $10^{-6}$ & $10^{-6}$ & $5\!\times\!10^{-6}$ & $10^{-6}$ & $2.5\!\times\!10^{-6}$ & $2.5\!\times\!10^{-6}$ \\
Weight decay & 0.01 & 0 & 0.01 & 0.01 & 0.01 & 0.01 \\
Batch & 32 & $128\!\times\!8$ & 32 & $16\!\times\!5$ & 112 & $14\!\times\!8$ \\
Max tokens & 16,384 & 8,192/512 & 16,384 & 8,192/8,192 & 8,320 & 19,872/128 \\
KL coefficient & -- & 0 & -- & 0.01 & -- & 0.04 \\
\bottomrule
\end{tabular}
\end{table}

\paragraph{Agentic.} SFT targets are built by UI-MOPD's released converter from each demonstration step's rationale, action description, and tool call~\citep{ui_mopd}. The RL teacher uses UI-MOPD's released GUI-action reward: $+1$ if the parsed tool call matches the reference action type and its arguments (e.g., a coordinate inside the annotated element box or case-insensitive text), $-0.5$ for any other parsable tool call, and $-1$ if no parsable tool call is produced. Each generation batch samples eight responses for each of 384 prompts; prompts whose responses include both $+1$ and lower rewards are retained until 128 fill an optimizer batch.

\paragraph{Reasoning and Perception.} The Reasoning SFT teacher uses NVIDIA's released chain-of-thought demonstrations~\citep{nvidia_geo3k}, whose final answers all match the Geo3K labels; the three demonstrations longer than 16,384 tokens are excluded, leaving 2,075. The Reasoning RL teacher draws its prompts from the same Geo3K training problems, and its reward is 1 if the final option letter parsed from the response matches the original Geo3K label and 0 otherwise, including responses without a parsable final answer. Each iteration samples five responses for each of 64 prompts and performs four updates of 16 prompts; the learning rate, batch sizes, sample count, and KL setting follow the Qwen3.5 Geo3K GRPO example in verl~\citep{sheng2024hybridflow}. Perception SFT targets a click at the center of the ground-truth box in 0--1000 coordinates, and the Perception RL reward is 1 if the coordinate in the response's tool call lies inside that box and 0 otherwise, including unparsable responses.

\subsection{Student Training Configurations}

Table~\ref{tab:training-configurations} reports the shared OPD settings for both student groups in each domain. Input batch counts prompts, and $n$ is the number of sampled responses per prompt. The resulting optimizer batches contain 1,024, 32, and 112 sequences for Agentic, Reasoning, and Perception, respectively.

\begin{table}[htbp]
\centering
\small
\setlength{\tabcolsep}{10pt}
\caption{Student OPD training settings.}
\label{tab:training-configurations}
\begin{tabular}{@{}lccc@{}}
\toprule
Parameter & Agentic & Reasoning & Perception \\
\midrule
Optimizer & AdamW & AdamW & AdamW \\
Learning rate & $10^{-6}$ & $1.875\!\times\!10^{-6}$ & $1.875\!\times\!10^{-6}$ \\
Weight decay & 0 & 0.01 & 0.01 \\
Input batch & 128 & 32 & 112 \\
Responses per prompt ($n$) & 8 & 1 & 1 \\
Warmup (updates) & 0 & 3 & 3 \\
\midrule
Max response tokens & 512 & 8,192 & 128 \\
Temperature & 1.0 & 1.0 & 0.9 \\
Top-$p$ & 1.0 & 0.95 & 0.9 \\
Sampling top-$k$ & Disabled & 20 & 50 \\
\bottomrule
\end{tabular}
\end{table}

All student runs use optimizer betas $(0.9,0.999)$, a constant learning rate after warmup, a gradient clipping norm of 1, and a maximum prompt length of 8,192 tokens. OPD uses no LoRA. Reasoning freezes the visual encoder and merger; Agentic and Perception freeze no parameters.

The sampling settings in Table~\ref{tab:training-configurations} apply to training rollouts. All teachers remain fixed during OPD. Agentic follows the batch size, responses per prompt, sequence limits, learning rate, and rollout sampling settings of UI-MOPD's student training~\citep{ui_mopd} and uses sampled-token distillation; Reasoning and Perception use full-vocabulary reverse KL. Each domain uses the same objective for both student groups.

\subsection{Checkpoint Selection and Evaluation}

Teacher checkpoints are selected to have comparable MobileWorld mean success rates for Agentic, Geo3K accuracies for Reasoning, and unweighted mean accuracies across six benchmarks for Perception. Student checkpoints are evaluated with the same procedures, and Appendix~\ref{app:checkpoint-results} reports all their benchmark results.

MobileWorld uses greedy decoding, at most 50 interaction steps, and up to four previous screenshots plus the current screenshot and action-history text.

Reasoning evaluates 601 Geo3K test problems, 1,000 MathVista testmini problems, 304 MathVision testmini problems, and 788 MathVerse testmini Vision-Only problems. Teachers and students generate one answer per problem with thinking disabled, temperature 1, top-$p$ 0.95, top-$k$ 20, presence penalty 1.5, and at most 8,192 output tokens in a 16,384-token context. We use the benchmark answer-scoring procedures; MathVista and MathVerse use Qwen3.5-122B-A10B for answer extraction or equivalence checking.

Perception evaluates ScreenSpot-v2 (1,272 examples), ScreenSpot-Pro (1,581), UI-Vision element grounding (5,479), MMBench-GUI L2 (3,594), OSWorld-G (564), and OSWorld-G-Refine (564). ScreenSpot-Pro uses the default configuration of its official evaluation repository. For ScreenSpot-v2, UI-Vision, MMBench-GUI L2, and both OSWorld-G variants, decoding uses temperature 0.7, top-$p$ 0.8, top-$k$ 20, and at most 4,096 output tokens. MMBench-GUI L2 averages its Basic and Advanced accuracies equally. Cross-benchmark means weight each benchmark equally.
\FloatBarrier

\section{Results Across Checkpoints}
\label{app:checkpoint-results}

Tables~\ref{tab:agentic-checkpoints}--\ref{tab:perception-checkpoints} report the benchmark results of the student checkpoints shown in Figure~\ref{fig:single-teacher-results}, together with those of the base model and both teachers.

\begin{table}[!htbp]
\centering
\footnotesize
\renewcommand{\arraystretch}{0.95}
\caption{MobileWorld results across OPD checkpoints. R1--R3: successes per round; Mean: success rate (\%).}
\label{tab:agentic-checkpoints}
\setlength{\tabcolsep}{4pt}
\begin{tabular}{@{}lcccccccc@{}}
\toprule
 & \multicolumn{4}{c}{SFT} & \multicolumn{4}{c}{RL} \\
\cmidrule(lr){2-5}\cmidrule(l){6-9}
Checkpoint & R1 & R2 & R3 & Mean & R1 & R2 & R3 & Mean \\
\midrule
Teacher & 28 & 23 & 29 & 22.79 & 26 & 28 & 28 & 23.36 \\
\midrule
Qwen3.5-9B & 22 & 20 & 20 & 17.66 & 22 & 20 & 20 & 17.66 \\
Step 8 & 21 & 28 & 21 & 19.94 & 17 & 23 & 25 & 18.52 \\
Step 16 & 19 & 19 & 20 & 16.52 & 25 & 18 & 23 & 18.80 \\
Step 24 & 23 & 22 & 22 & 19.09 & 23 & 29 & 24 & 21.65 \\
Step 32 & 23 & 22 & 24 & 19.66 & 33 & 26 & 26 & 24.22 \\
Step 40 & 19 & 24 & 21 & 18.23 & 30 & 29 & 25 & 23.93 \\
Step 48 & 21 & 24 & 24 & 19.66 & 25 & 28 & 30 & 23.65 \\
Step 56 & 22 & 22 & 21 & 18.52 & 31 & 29 & 24 & 23.93 \\
Step 64 & 18 & 18 & 22 & 16.52 & 27 & 27 & 25 & 22.51 \\
\bottomrule
\end{tabular}
\end{table}

\begin{table}[!htbp]
\centering
\footnotesize
\renewcommand{\arraystretch}{0.95}
\caption{Reasoning accuracy (\%) across OPD checkpoints.}
\label{tab:reasoning-checkpoints}
\setlength{\tabcolsep}{4pt}
\begin{tabular}{@{}lcccccc@{}}
\toprule
Model & Step & Geo3K & MathVista & MathVision & MathVerse & Avg. \\
\midrule
Qwen3.5-9B & -- & 79.70 & 84.00 & 57.24 & 70.18 & 72.78 \\
SFT teacher & -- & 88.52 & 81.70 & 52.63 & 67.13 & 72.50 \\
RL teacher & -- & 84.19 & 83.60 & 57.57 & 73.10 & 74.61 \\
\midrule
Student $\leftarrow$ SFT & 4 & 85.86 & 83.70 & 51.97 & 70.18 & 72.93 \\
 & 8 & 88.85 & 82.80 & 54.28 & 66.50 & 73.11 \\
 & 12 & 90.68 & 82.90 & 53.95 & 67.77 & 73.82 \\
 & 16 & 87.35 & 81.70 & 58.55 & 70.69 & 74.57 \\
 & 20 & 85.19 & 82.10 & 56.25 & 68.53 & 73.02 \\
 & 24 & 68.72 & 81.90 & 59.54 & 70.56 & 70.18 \\
 & 28 & 68.89 & 81.90 & 58.22 & 71.57 & 70.15 \\
 & 32 & 72.88 & 81.90 & 58.88 & 70.94 & 71.15 \\
\midrule
Student $\leftarrow$ RL & 4 & 82.03 & 82.10 & 56.58 & 70.05 & 72.69 \\
 & 8 & 90.52 & 83.70 & 57.89 & 72.46 & 76.14 \\
 & 12 & 89.85 & 84.30 & 58.22 & 74.37 & 76.68 \\
 & 16 & 90.18 & 82.70 & 57.24 & 71.57 & 75.42 \\
 & 20 & 91.51 & 83.10 & 58.55 & 71.07 & 76.06 \\
 & 24 & 90.68 & 83.60 & 60.53 & 70.30 & 76.28 \\
 & 28 & 89.68 & 83.90 & 57.24 & 70.81 & 75.41 \\
 & 32 & 90.35 & 84.10 & 57.24 & 71.83 & 75.88 \\
\bottomrule
\end{tabular}
\end{table}

\begin{table}[!htbp]
\centering
\footnotesize
\renewcommand{\arraystretch}{0.95}
\caption{Perception accuracy (\%) across OPD checkpoints.}
\label{tab:perception-checkpoints}
\setlength{\tabcolsep}{4pt}
\begin{tabular}{@{}lcccccccc@{}}
\toprule
Model & Step & \shortstack{ScreenSpot-\\v2} & \shortstack{ScreenSpot-\\Pro} & UI-Vision & \shortstack{MMBench-\\GUI L2} & \shortstack{OSWorld-\\G} & \shortstack{OSWorld-\\G-R} & Avg. \\
\midrule
Qwen3.5-9B & -- & 91.75 & 63.00 & 26.92 & 80.41 & 61.35 & 67.73 & 65.19 \\
SFT teacher & -- & 92.92 & 65.15 & 33.67 & 83.87 & 62.94 & 70.92 & 68.24 \\
RL teacher & -- & 93.16 & 65.28 & 30.52 & 82.39 & 63.65 & 71.28 & 67.71 \\
\midrule
Student $\leftarrow$ SFT & 2 & 91.12 & 62.87 & 26.14 & 79.53 & 61.52 & 68.44 & 64.94 \\
 & 4 & 91.35 & 63.19 & 27.52 & 81.17 & 61.52 & 68.44 & 65.53 \\
 & 6 & 92.61 & 63.38 & 32.94 & 82.66 & 63.30 & 67.91 & 67.13 \\
 & 8 & 93.24 & 63.76 & 38.38 & 84.32 & 65.60 & 71.10 & 69.40 \\
 & 10 & 94.50 & 64.20 & 39.08 & 84.54 & 64.72 & 71.28 & 69.72 \\
 & 20 & 94.18 & 64.83 & 40.28 & 85.28 & 65.78 & 71.81 & 70.36 \\
 & 30 & 94.26 & 64.64 & 38.88 & 84.73 & 64.18 & 73.23 & 69.99 \\
 & 40 & 93.63 & 64.96 & 39.04 & 85.12 & 65.60 & 71.63 & 70.00 \\
 & 50 & 93.95 & 65.02 & 38.95 & 84.76 & 65.78 & 71.63 & 70.02 \\
 & 60 & 94.34 & 64.20 & 37.32 & 84.41 & 64.01 & 72.16 & 69.41 \\
 & 62 & 93.47 & 64.07 & 38.55 & 84.19 & 61.52 & 71.45 & 68.88 \\
\midrule
Student $\leftarrow$ RL & 2 & 91.43 & 62.49 & 26.26 & 79.91 & 58.51 & 65.60 & 64.03 \\
 & 4 & 92.69 & 63.44 & 28.47 & 81.09 & 61.35 & 67.91 & 65.83 \\
 & 6 & 92.77 & 65.34 & 33.51 & 83.81 & 63.12 & 70.74 & 68.21 \\
 & 8 & 93.63 & 65.97 & 40.70 & 86.03 & 67.02 & 72.16 & 70.92 \\
 & 10 & 94.89 & 65.97 & 42.31 & 86.54 & 65.60 & 71.99 & 71.22 \\
 & 20 & 93.71 & 66.67 & 41.74 & 86.70 & 66.84 & 70.21 & 70.98 \\
 & 30 & 94.03 & 66.10 & 38.09 & 85.26 & 63.48 & 70.74 & 69.62 \\
 & 40 & 92.85 & 66.98 & 35.39 & 85.20 & 63.30 & 70.39 & 69.02 \\
 & 50 & 93.71 & 66.54 & 35.97 & 84.70 & 63.30 & 71.45 & 69.28 \\
 & 60 & 93.40 & 67.81 & 37.32 & 85.12 & 64.36 & 69.33 & 69.56 \\
 & 62 & 93.63 & 66.60 & 35.48 & 84.65 & 64.36 & 69.50 & 69.04 \\
\bottomrule
\end{tabular}
\end{table}
\FloatBarrier

Table~\ref{tab:mopd-checkpoints} reports all MOPD student checkpoints shown in Figure~\ref{fig:mopd-training-dynamics}.

\begin{table}[!htbp]
\centering
\small
\caption{MOPD performance (\%) across checkpoints. Bold marks each group's best three-domain mean.}
\label{tab:mopd-checkpoints}
\begin{tabular*}{\linewidth}{@{\extracolsep{\fill}}lccccc@{}}
\toprule
Teacher group & Step & MobileWorld & Geo3K & Perception & Mean \\
\midrule
SFT & 8 & 17.09 & 88.52 & 64.19 & 56.60 \\
 & 16 & 20.23 & 88.52 & 65.24 & 57.99 \\
 & 24 & 18.52 & 89.18 & 66.29 & 58.00 \\
 & 32 & 19.37 & 90.18 & 67.09 & 58.88 \\
 & 40 & 17.95 & 89.68 & 67.64 & 58.42 \\
 & 48 & 20.23 & 89.18 & 67.10 & 58.84 \\
 & \textbf{51} & 19.94 & 89.68 & 67.30 & \textbf{58.97} \\
\midrule
RL & 8 & 19.66 & 85.69 & 65.55 & 56.97 \\
 & 16 & 21.65 & 86.86 & 68.52 & 59.01 \\
 & 24 & 21.94 & 87.69 & 69.28 & 59.63 \\
 & 32 & 22.22 & 86.69 & 69.67 & 59.53 \\
 & \textbf{40} & 23.93 & 86.52 & 69.68 & \textbf{60.04} \\
 & 48 & 22.79 & 85.36 & 69.91 & 59.35 \\
 & 51 & 20.80 & 85.52 & 69.61 & 58.65 \\
\bottomrule
\end{tabular*}
\end{table}
\FloatBarrier

\section{Further Directions}
\label{app:further-directions}

Building on our MOPD results, future work could examine how the choice of teacher-training algorithm interacts with domain mixing and teacher routing~\citep{ma2026mopd,men2026physics,ui_mopd}. Our MobileWorld results further suggest that the choice of teacher-training algorithm may be particularly important for OPD in Agentic tasks. A valuable next step is to test whether this effect extends to more challenging agentic benchmarks, including long-horizon settings such as Long-Horizon-Terminal-Bench~\citep{li2026long}. Data synthesis methods that improve post-training data diversity~\citep{li2026less} could further support studying how this effect varies across training distributions.

\end{document}